\documentclass[10pt,conference,final,twocolumn]{IEEEtran}
\usepackage{amsmath,amssymb,graphicx,xcolor}
\usepackage{subfigure,algorithm,algpseudocode,subfig,multirow,subfloat}
\usepackage{epsfig,epstopdf}

\definecolor{myred}{rgb}{0.7, 0, 0}
\newcommand{\mb}[1]{\mathbf{#1}}
\newcommand{\bs}[1]{\boldsymbol{#1}}

\begin{document}
\title{A New Clustering-Based Technique for the Acceleration of Deep Convolutional Networks}

\author{
\IEEEauthorblockN{Erion Vasilis Pikoulis$^{1,2}$, Christos Mavrokefalidis$^{1,2}$, Aris S. Lalos$^{1}$}
\IEEEauthorblockA{$^1$Industrial Systems Institute, Patras Science Park, Greece\\
$^2$Computer Engineering and Informatics Dept., University of Patras, Greece\\
Emails: \{pikoulis,maurokef\}@ceid.upatras.gr, lalos@isi.gr
}
}

\maketitle

\begin{abstract}
Deep learning and especially the use of Deep Neural Networks (DNNs) provides impressive results in various regression and classification tasks. However, to achieve these results, there is a high demand for computing and storing resources. This becomes problematic when, for instance, real-time, mobile applications are considered, in which the involved (embedded) devices have limited resources. A common way of addressing this problem is to transform the original large pre-trained networks into  new  smaller models, by utilizing  Model Compression  and  Acceleration  (MCA)  techniques. Within the MCA framework, we propose a clustering-based approach that is able to increase the number of employed centroids/representatives, while at the same time, have an acceleration gain compared to conventional, $k$-means based approaches. This is achieved by imposing a special structure to the employed representatives, which is enabled by the particularities of the problem at hand. Moreover, the theoretical acceleration gains are presented and the key system hyper-parameters that affect that gain, are identified. Extensive evaluation studies carried out using various state-of-the-art DNN models trained in image classification, validate the superiority of the proposed method as compared for its use in MCA tasks.
\end{abstract}

\section{Introduction}

Deep Neural Networks (DNN) \cite{Goodfellow2016} have emerged recently as a central ingredient in 
many modern artificial intelligence applications \cite{Najafabadi2015,Cao2018,Wang2018,Suzuki2017,Bolton2019}. 
However, the impressive performance that has been reported in the literature, is closely related to the size of the DNNs, which, for state-of-the-art models can reach to tens, or even hundreds of millions of parameters 
(e.g., 138 millions of parameters are used by the Visual Geometry Group (VGG) DNN \cite{Simonyan2014}). This leads to vast computing and storage requirements
during both the training phase of the DNNs and (most importantly) the operational (or inference) phase, i.e., when the DNNs are actually employed. 
In real applications, these requirements are usually tackled via high-performance computing platforms \cite{Hatcher2018} that include graphics processing units (GPUs). 

Nowadays, there is an increasing interest in blending deep learning and mobile computing in which platforms of limited resources are employed \cite{Tang2017}, including smart devices (such as phones, watches, and embedded sensors). In order for these platforms
 to exploit the DNN gains, especially, during 
the inference phase, two main lines of research can be identified. In the first one, new compact, smaller DNN models \cite{Deng2020} are designed or searched for in a design space for the applications at hand (e.g., SqueezeNet \cite{Iandola2016}, MobileNets \cite{Howard2017}, and EfficientNet \cite{Tan2019}). In the second line, existing pre-trained, highly performing DNN models 
(e.g., AlexNet \cite{krizhevsky2012}, VGG \cite{Simonyan2014}, Residual Net (ResNet) \cite{He2016}, and many more) are transformed into new smaller models by utilizing Model Compression and Acceleration (MCA) techniques \cite{Deng2020}, \cite{Sze2017}, \cite{Cheng2018}, \cite{Zhang2019}. 
The importance of this line of research stems from the fact that, apart from their standalone use, state-of-the-art, pre-trained DNN models can also be utilized as back-bone modules in models designed for different (but similar in nature) applications. For example, the convolutional layers of AlexNet and VGG (that are originally trained for image classification tasks), constitute the core modules of the R-CNN \cite{Girshick2014} and Fast R-CNN \cite{Girshick2015}, respectively, object detectors. 



The MCA-related literature has been increasing in recent years and there are numerous surveys that provide a comprehensive overview of
the area (\cite{Deng2020}, \cite{Sze2017}, \cite{Cheng2018}, \cite{Zhang2019}). Roughly speaking (and by no means being exhaustive), 
some of the earliest works proposed parameter pruning, in which, unimportant parameters (e.g., filters \cite{Li2016}, \cite{Lin2019}) are removed and, hence, not considered during the inference phase of the DNN deployment. Other works focus on limiting the representation of the involved parameter by reducing their bit-width or 
increasing common representations via the design of codebooks 
(e.g., scalar \cite{Gupta2015}, vector and product quantization \cite{Gong2014}). Finally,
several works employ tensor / matrix decompositions on the involved quantities (e.g., filters) into factors by utilizing, for instance, 
low-rankness \cite{Bhattacharya2016}.

In this paper, a new MCA technique for pre-trained DNNs is described and evaluated. The highlights of the paper are outlined as follows:
\begin{itemize}
    \item We propose a novel codebook design procedure that, for the same target acceleration, leads to larger codebooks than the typically used $k$-means-based approaches, thus improving considerably the quantization error.
    \item This is achieved by imposing a special structure to the learned codewords based on a Dictionary Learning framework.
    \item Theoretical analysis is provided for determining the parameters that dictate the structure of the codebook.
    \item The efficacy of the proposed MCA technique is assessed on three state-of-the-art DNN models (VGG, ResNet, SqueezeNet) 
    on the demanding ILSVRC2012 dataset \cite{krizhevsky2012}, achieving up to $100\%$ (or $2\times$) acceleration gain over the conventional approach, for the same quantization error.
    \item The application of the proposed technique for the selected DNNs results in significantly accelerated models with limited 
    performance degradation.
\end{itemize}


In the following, first, the relevant bibliography is presented and the positioning of this paper is 
described. Then, in Sec.~\ref{sec:problem}, the problem is formulated, while, in Sec.~\ref{sec:proposed}, the proposed technique is explained. 
Simulation results are presented in Sec.~\ref{sec:simulations}. Finally, Sec.~\ref{sec:conclusions} concludes the paper.

\section{Relevant bibliography}

The work in this paper is related to MCA techniques that utilize a codebook for 
quantizing network parameters. The codebook contains representative quantities (called codewords, centroids, etc.) 
for the parameters to be approximated (e.g., vectors of weights). As multiple network parameters are mapped to a single
representative quantity, such approaches are also called parameter sharing techniques. 

Towards this end, \cite{Gong2014} proposed vector quantization methods (using the $k$-means algorithm) for 
estimating the desired codebook with the aim to compress the weights of fully connected layers. 
In \cite{Han2015}, a three stage method was presented in which
parameter pruning was followed by a codebook design for parameter quantization and concluded with Huffman coding. 
In \cite{Cheng2018_quantized}, a new codebook design was devised that improved upon or generalized works like \cite{Gong2014}, \cite{Han2015}. In \cite{Cheng2018_quantized}, the proposed approach could be applied to 
both convolutional and fully connected layers, while two
cost-functions for estimating the involved codebooks, were devised that minimized the quantization error of the weights and the representation error of the layers' output (namely, the error between the outputs of the original and the accelerated layers for a given input), respectively. In \cite{Son2018}, the proposed technique operates on scaled versions
of the 2D kernels for estimating the desired centroids using the $k$-means algorithm. 
Then, during fine-tuning, both the scales and the centroids are considered free
variables to be updated. In \cite{Wu2018}, \cite{Hu2019} and \cite{Ben2019}, proper regularization terms are introduced and via 
re-training procedures the resulted weights can be more easily clustered using the $k$-means algorithm. The work in \cite{Stock2019} adopts product quantization and focuses on the representation error of the layer outputs. Finally, in \cite{Dupuis2020}, a
methodology is devised for determining the size of the codebooks by introducing a sensitivity analysis per layer in order
to assess the impact of compression on the accuracy performance.

The proposed MCA technique is mostly related to works like \cite{Gong2014}, \cite{Han2015}, and \cite{Cheng2018_quantized}. Here, however,
by exploiting the special structure of the weights to be quantized, improved quantization error is achieved. For assessing the performance, \cite{Cheng2018_quantized} is selected as a baseline. Moreover, although, here, a new MCA technique is outlined, the proposed dictionary-learning
approach for designing the codebook, can be actually utilized by all the works mentioned above (i.e., instead of the commonly used $k$-means 
algorithm).

\begin{figure}[t]
\centering
\includegraphics[scale=0.5]{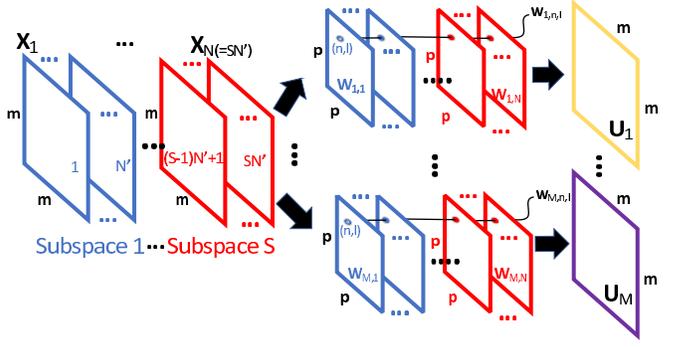}
\caption{The linear operation of a single convolutional layer.}
\label{fig:system}
\end{figure}
	
\section{Problem Formulation}\label{sec:problem}

The core operation performed by a convolutional layer and the involved quantities, are depicted in Fig.~\ref{fig:system}. In particular, the input volume
consists of $N$ channels $\mathbf{X}_i$, $i=1,2,\ldots,N$. Also, there are $M$ kernel volumes and the $k$-th kernel volume has $N$ filters  $\mathbf{W}_{k,i}$, $k=1,2,\ldots,M$, $i=1,2,\dots,N$. For simplicity, it is assumed that the dimensions of the $\mathbf{X}_i$'s,  $\mathbf{W}_{k,i}$'s and $\mathbf{U}_{k}$'s are $m\times m$, $p\times p$, and $m\times m$, respectively.

The convolution of the input volume with the $k$-th kernel volume
is given by 
\begin{equation}
    \mathbf{U}_{k}=\sum_{i=1}^N\mathbf{X}_i\star \mathbf{W}_{k,i},
    \label{eq:uk}
\end{equation}
where $\star$ denotes the 2D convolution operation.

In order to proceed and describe the entities to be clustered (i.e., coded by the codebook that will be designed), 
\eqref{eq:uk} is re-written in order to describe the $(i,j)$-th element $\mathbf{U}_{k}[i,j]$ as
\begin{equation}
    \mathbf{U}_{k}[i,j]=\sum_{n,l\in \mathcal{R}_{i,j}} \mathbf{x}^\textrm{T}_{n,l}\mathbf{w}_{k,i-n,j-l},
    \label{eq:ukij}
\end{equation}
where $\mathbf{x}_{i,j}=[\mathbf{X}_1[i,j],\dots,\mathbf{X}_N[i,j]]^\textrm{T}$ contains the samples at the $(i,j)$-th position of all 
input channels. Also, $\mathbf{w}_{k,u,v}=[\mathbf{W}_{k,1}[u,v],\dots,\mathbf{W}_{k,N}[u,v]]^\textrm{T}$ contains the filter weights at the $(u,v)$ 
position of all channels in the $k$-th kernel volume. Finally, the set $\mathcal{R}_{i,j}$ contains $p^2$ indices around the position $(i,j)$.

In the product quantization framework, the $N$-dimensional vector space is partitioned into $S$, $N'$-dimensional subspaces with $N'=N/S$, so that the $s$-th subspace spans dimensions 
$[(s-1)N'+1,\dots,sN']$, $s=1,\dots,S$. Let us now partition vectors $\mathbf{x}_{i,j}$, $\mathbf{w}_{k,u,v}$ defined in Eq. 
\eqref{eq:ukij}, accordingly, as
\begin{IEEEeqnarray}{rcl}
\mathbf{x}_{i,j}&=&[(\mathbf{x}^1_{i,j})^\textrm{T},\dots,(\mathbf{x}^S_{i,j})^\textrm{T}]^\textrm{T}, \label{eq.:x_ij}\\    
\mathbf{w}_{k,u,v}&=&[(\mathbf{w}^1_{k,u,v})^\textrm{T},\dots,(\mathbf{w}^S_{k,u,v})^\textrm{T}]^\textrm{T}, \label{eq.:w_knl}
\end{IEEEeqnarray}
where each of the sub-vectors lies in $N'$-D space. Then, \eqref{eq:ukij} can be rewritten as
\begin{equation}
    \mathbf{U}_{k}[i,j]=\sum_{s=1}^S\sum_{n,l\in \mathcal{R}_{i,j}} (\mathbf{x}^s_{n,l})^\textrm{T}\mathbf{w}^s_{k,i-n,j-l},
    \label{eq:ukij_3}
\end{equation}
where the inner sum denotes the contribution of the $s$-th subspace to the $k$-th convolutional output, at position $(i,j)$. 


For each subspace, the goal of product quantization is to perform vector quantization to the $Mp^2$ kernel sub-vectors lying in $s$-th subspace, and cluster them into $K_s\ll Mp^2$ clusters. This way, each sub-vector is represented by the centroid of the cluster it belongs to, reducing accordingly the number of required dot-products. To be more specific, the acceleration occurs because the original dot-products between the input and the $Mp^2$ kernel sub-vectors, are approximated by the ones between the input and the $K_s$ centroids/representatives. 

\section{Dictionary-learning-based weight clustering}
\label{sec:proposed}

In this section, first, the proposed codebook structure for approximating the kernel sub-vectors, is described and discussed in comparison with the conventional codebook structure
that appears in current literature. This discussion is also extended towards the gains that are achieved through a computational complexity analysis. Then, the proposed codebook design is approached as a Dictionary Learning (DL) problem, which actually treats the $k$-means-based conventional
codebook design as a special case. The latter will be referred to as Vector Quantization (VQ) in the following. Finally, some implementation details are described concerning the initialization
of the involved parameters when applying the proposed DL solution.


\subsection{Proposed approximation}
Let us first define the kernel approximation scheme incurred by the conventional codebook structure, as follows:
\begin{equation}
\mathbf{W}\approx \mathbf{C}\mathbf{\Gamma},
\label{eq:W_approx_CG}
\end{equation}
where the columns of $\mb{W}\in\mathbb{R}^{N'\times p^2M}$, $\mb{C}\in\mathbb{R}^{N'\times K_{vq}}$, and $\mb{\Gamma}\in\mathbb{R}^{K_{vq}\times p^2M}$, contain the kernel sub-vectors (of a particular subspace), the representatives (or cluster centroids), and assignment vectors, respectively. Specifically, each column of $\mb{\Gamma}$ has exactly one non-zero element, equal to $1$, meaning that each column of $\mb{W}$ is approximated by one column of $\mb{C}$. Thus, in the conventional case, the $Mp^2$ sub-vectors are approximated by $K_{vq}\ll p^2M$ representatives, using the codebook $\mb{C}$.

Instead, in this paper, the following approximation is proposed: 
\begin{equation}
\mb{W}\approx\mb{D}\mb{\Lambda}\mb{\Gamma},
\label{eq:W_approx_DLG}
\end{equation}
where $\mb{W}\in\mathbb{R}^{N'\times p^2M}$ and $\mb{\Gamma}\in\mathbb{R}^{K_{dl}\times p^2M}$ are defined as in \eqref{eq:W_approx_CG}, while $\mb{D}\in\mathbb{R}^{N'\times L_{dl}}$ and $\mb{\Lambda}\in\mathbb{R}^{L_{dl}\times K_{dl}}$ denote the dictionary and the matrix of sparse coefficients, respectively. Specifically, the columns of $\mb{D}$ (called dictionary atoms), are normalized, while $\mb{\Lambda}$ is a sparse matrix in the sense that each of its columns contains at most $\alpha$ non-zero elements, with $\alpha$ being the sparsity level. Thus, under the proposed scheme, the $p^2M$ sub-vectors are approximated via $K_{dl}$ representatives contained in the codebook $\mb{D}\mb{\Lambda}$. In turn, these representatives are obtained as linear combinations of at most $\alpha$ atoms from a dictionary of size $L_{dl}$, with $L_{dl}<K_{dl}\ll p^2M$. Note that the matrix approximation defined in \eqref{eq:W_approx_DLG} can be viewed as a special case of the general Dictionary Learning (DL) problem \cite{Dumitrescu2018}, which is why we call our acceleration technique as a DL-based one. 



It should be noted that, in the general case, the proposed approximation requires more representatives than the conventional approach (i.e., $K_{dl}>K_{vq}$), for achieving the same quantization (i.e. approximation) error. This is by definition since the conventional codebook $\mb{C}$ is obtained in an unconstrained fashion, while the proposed codebook $\mb{D}\mb{\Lambda}$ follows a specific structure. Although it seems counter intuitive (in the sense that the proposed approximation is less efficient than the conventional one, in the general case), due to the particularities of the problem at hand, namely, due to the fact that the ``data points'' in $\mb{W}$ are in fact filters used in convolution operations, the proposed approximation results actually in significantly higher acceleration ratios for the same quantization error, as it is going to be demonstrated. This is because, due to the linearity of the operations performed in the convolutional layer, the sparse coefficients in $\mb{\Lambda}$ need only be applied to the convolution between the input and the dictionary atoms in $\mb{D}$, instead of the atoms themselves. This endows the proposed approximation scheme with the flexibility to use a number of representatives $K_{dl}$ that is several times larger than $K_{vq}$, while restricting the size of the dictionary (so that $L_{dl}\ll K_{vq}$) thus reducing the number of ``heavy'' convolutions, as it will become clearer in the following subsection.


%
\subsection{Computational complexity analysis}

Since the core operations of a DNN are ultimately translated into dot-products between input and kernel vectors, the computational complexity of 
a DNN is usually measured in terms of the number of Multiply and Accumulate (MAC) operations. A MAC is dominated by the involved multiplication (MUL), which is a significantly ``heavier'' computation than the involved addition. As such, in the subsequent analysis, in order to 
compare the techniques on a common ground, MAC and MUL operations are going to be used interchangeably, i.e, a MAC will be considered 
equivalent to one MUL, so that the computational cost is measured as the number of MULs. 

Let us first begin by examining the computational complexity of the original layer, where, by arranging the $m^2$ input sub-vectors of the $s$-th subspace in the columns of a matrix $\mb{X}\in\mathbb{R}^{N'\times m^2}$, we see that the convolution operation involves the calculation of a matrix product of the form:
\begin{equation}
    \mb{Y}=\mb{X}^{\textrm{T}}\mb{W},
    \label{eq:Y=X'W}
\end{equation}
where $\mb{W}$ contains the kernel sub-vectors (as defined in \eqref{eq:W_approx_CG}), followed by the appropriate summation of the dot-products according to \eqref{eq:ukij_3}. Since calculating $\mb{Y}$ requires $m^2p^2MN'$ multiplications, the overall (i.e. for all $S$ subspaces) computational complexity of the original convolutional layer, measured in MULs, is obtained as:
\begin{equation}
    \label{eq:T(F)_origt}
    \mathcal{T}_o=m^2p^2M(SN')=m^2p^2MN.   
\end{equation}
In the VQ case (described by the approximation in \eqref{eq:W_approx_CG}), the approximate $\mb{Y}$ is obtained as:
\begin{equation}
   \mb{Y}\approx (\mb{X}^{\textrm{T}}\mb{C})\mb{\Gamma},
   \label{eq:Y=XCG}
\end{equation}
namely, it involves calculating the dot-products between input sub-vectors and representatives and then ``plugging'' the results appropriately, according to the columns of $\mb{\Gamma}$. Calculating $\mb{X}^{\textrm{T}}\mb{C}$ requires only $m^2N'K_{vq}$ MULs (as $k_{vq}\ll Mp^2$), 
meaning that the overall computational complexity for the approximate convolutional output is reduced to:
\begin{equation}
    \label{eq:T(F)_vq}
    \mathcal{T}_{vq}(K_{vq})=m^2(SN')K_{vq}=m^2NK_{vq}.
\end{equation}
Finally, for the DL-based approximation scheme, we can write:
\begin{equation}
   \mb{Y}\approx \left((\mb{X}^{\textrm{T}}\mb{D})\mb{\Lambda}\right)\mb{\Gamma},
   \label{eq:Y=XDLG}
\end{equation}
meaning that, in this case, calculating the approximate $\mb{Y}$ is a two-stage operation. First, we calculate $\mb{X}^{\textrm{T}}\mb{D}$, i.e., the dot-products between the input and the dictionary atoms, which requires $m^2N'L_{dl}$ MULs, where $L_{dl}$ denotes the dictionary size. Subsequently, the results are combined according to the columns of $\mb{\Lambda}$, which requires $\alpha\, m^2 K_{dl}$ additional MULs
Thus, the overall computational complexity for the approximate convolutional output in the DL case, is obtained as: 
\begin{equation}
    \label{eq:T(F)_dl}
    \mathcal{T}_{dl}(K_{dl},L_{dl},\alpha)=m^2(NL_{dl}+\alpha S K_{dl}).    
\end{equation}

Accordingly, the acceleration ratio (namely, the ratio of original vs accelerated computational complexities) achieved by the two rival approaches, can be written as follows:
\begin{IEEEeqnarray}{rl}
\label{eq:accel_vq}
\rho_{vq}&\equiv \frac{\mathcal{T}_o}{\mathcal{T}_{vq}}=\frac{p^2M}{K_{vq}}\\
\label{eq:accel_dl}
\rho_{dl}&\equiv \frac{\mathcal{T}_o}{\mathcal{T}_{dl}}=\frac{p^2M}{L_{dl}+\displaystyle{\frac{\alpha}{N'}}K_{dl}}.
\end{IEEEeqnarray}

Of great interest is also the relative acceleration between the proposed  
DL-based and the VQ approach, which will also provide rules for selecting the free parameters of the proposed technique. First, in order to have a better representation error, we set the number of representatives used by the proposed technique as a multiple of the representatives used by the VQ approach, i.e., $K_{dl}=c\,K_{vq}$, $c>1$. Then, using \eqref{eq:T(F)_vq}, \eqref{eq:T(F)_dl}, we can see that for the DL-based approximation to achieve at least the same acceleration with the VQ technique, i.e., for    $\mathcal{T}_{dl}(K_{dl},L_{dl},\alpha)\leq\mathcal{T}_{vq}(K_{vq})$ to hold, the following inequality should hold regarding the size of the used dictionary:
\begin{equation}
    \label{eq:T_dl vs T_vq}
    L_{dl}\leq K_{vq}\left(1-\frac{\alpha\,c}{N'}\right).
\end{equation}
As we are going to demonstrate in our experimental results for various combinations of the coefficient $c$ and sparsity level $\alpha$, and for \eqref{eq:T_dl vs T_vq} holding with equality (i.e., for the two rivals achieving the same acceleration ratio), the proposed technique leads to a significantly better approximation of the original weights, which ultimately translates into better classification accuracy for the accelerated DNNs.

\subsection{Proposed algorithm}

For deriving the matrix factorizations described by the proposed weight factorization in \eqref{eq:W_approx_DLG}, the quantization error between the original $\mb{W}$ and its approximate version is minimized. In particular,
the following minimization problem is defined. 
\begin{IEEEeqnarray}{rl}
\min_{\mb{D},\mb{\Lambda},\mb{\Gamma}} &\quad ||\mb{W}-\mb{D}\mb{\Lambda}\mb{\Gamma}||_F^2 \label{eq:ss_dl_dagnostic}\\
\nonumber
\text{s.t.}   &\quad ||\mb{d}_i||_2^2=1,\,i=1,\ldots,L_{dl},\\
            \nonumber
              &\quad ||\bs{\lambda}_i||_0\leq\alpha,\,i=1,\ldots,K_{dl},\\ 
            \nonumber &\quad||\bs{\gamma}_i||_0=1,\,\,\mb{1}^\textrm{T} \bs{\gamma}_i=1,\quad i=1\ldots Mp^2, 
\end{IEEEeqnarray}
where $||\cdot||_F$, $||\cdot||_2$, $||\cdot||_0$ denote the Frobenius, $l_2$, and $l_0$ norms, respectively, while the last constraint ensures that the elements of $\mb{\Gamma}$ take values in $\{0,1\}$ and each of its columns has exactly one non-zero element.

In order to solve \eqref{eq:ss_dl_dagnostic}, we follow a strategy of alternating optimizations over each set of parameters, leading to the following three sub-problems:\\
\paragraph{Sparse coding}
With $\mb{D}$, $\mb{\Gamma}$ fixed, the loss function in \eqref{eq:ss_dl_dagnostic} can be rewritten as follows:
\begin{equation}
    \bigg|\bigg|\mb{W}-\sum_{i=1}^{K_{dl}} (\mb{D}\bs{\lambda}_i)\widetilde{\bs{\gamma}}_i\bigg|\bigg|_F
    =\sum_{i=1}^{K_{dl}}\big|\big|\mb{W}_{I_i}-(\mb{D}\bs{\lambda}_i)\,\mb{1}^\textrm{T}\big|\big|_F,
    \label{eq:ss_dl_sparse_cod_dagnostic2}
\end{equation}
where $\mb{y}_i$, $\widetilde{\mb{y}}_i$ is used to denote the $i$-th column and row of matrix $\mb{Y}$, respectively, $I_i=\{j|\gamma_{ij}=1\}$ is the set of indices of the non-zero elements of $\widetilde{\bs{\gamma}}_i$, $\mb{W}_{I_i}$ is the submatrix formed by the columns of $\mb{W}$ indexed by $I_i$, while $\mb{1}$ denotes the all-ones vector of dimension $|I_i|$.
    
Observing \eqref{eq:ss_dl_sparse_cod_dagnostic2}, due to the $l_0$-norm constraints on $\mb{\Gamma}$, $I_i$'s, $i=1,\dots,K_{dl}$, are a partition of $\{1, 2, \ldots, p^2M\}$, meaning that the minimization of \eqref{eq:ss_dl_sparse_cod_dagnostic2} over $\mb{\Lambda}$ is translated into $K_{dl}$ separate sub-problems, one for each of $\mb{\Lambda}$'s columns:
\begin{IEEEeqnarray}{rl}
\label{eq:ss_dl_dagnostic_sparse_coding}
\min_{\bs{\zeta}} &\quad ||\mb{W}_{I_i}-(\mb{D}\bs{\zeta})\,\mb{1}^\textrm{T}||_F^2 \\
\nonumber
\text{s.t.}   &\quad ||\bs{\zeta}||_0\leq\alpha.
\end{IEEEeqnarray}
In order to solve \eqref{eq:ss_dl_dagnostic_sparse_coding}, we follow an Orthogonal Matching Pursuit (OMP) approach, which builds the support of the sparse representation (non-zero elements of $\bs{\lambda}_i$), by adding one dictionary atom at a time, up to
$\alpha$ atoms \cite{Dumitrescu2018}.

This sparse coding sub-problem is outlined in lines $5$-$12$ of Table \ref{table:algorithm}. There, $\mathcal{S}$ denotes a set of non-zero indices, while $\mb{D}_{\mathcal{S}}$ and $\bs{\zeta}_{\mathcal{S}}$ contain the columns of $\mb{D}$ and
the elements of $\bs{\zeta}$ indexed by $\mathcal{S}$, respectively. 

 \paragraph{Dictionary update} With $\mb{\Lambda}$, $\mb{\Gamma}$ fixed, we write the loss function in \eqref{eq:ss_dl_dagnostic} as follows:
\begin{equation}
    \mathcal{E}  =  ||\mb{W}-\mb{D}\mb{G}||_F^2=\bigg|\bigg|\mb{W}-\sum_{i=1}^{L_{dl}}\mb{d}_i \widetilde{\mb{g}}_i\bigg|\bigg|_F^2,
    \label{eq:E_W-Sdg}
\end{equation}
where $\widetilde{\mb{g}}_i$ denotes the $i$-th row of $\mb{G}=\mb{\Lambda}\mb{\Gamma}$. Thus, the dictionary update step translates to minimizing $\mathcal{E}$ under the $l_2$-norm constraint for the dictionary atoms. In order to solve this problem, we follow the coordinate-descent-based approach outlined in Algorithm 3.5 of \cite{Dumitrescu2018}. This sub-problem is described in lines $13$-$18$ of Table \ref{table:algorithm}.

\paragraph{Assignment update} With $\mb{D}$, $\mb{\Lambda}$ fixed, the loss function in \eqref{eq:ss_dl_dagnostic} takes the following form:
\begin{equation}
    \mathcal{E}=||\mb{W}-\widetilde{\mb{C}}\mb{\Gamma}||_F^2=\sum_{i=1}^{p^2M} ||\mb{w}_i-\widetilde{\mb{C}}\bs{\gamma}_i||_2^2
    \label{eq:E_Sw_Cg}
\end{equation}
where $\widetilde{\mb{C}}=\mb{D}\mb{\Lambda}$ is the  $N'\times K_{dl}$ matrix of representatives. Taking into account the special structure of $\mb{\Gamma}$, updating $\bs{\gamma}_i$ is equivalent to determining the position $j_i$ of its non-zero (unity) element, which 
simply assigns $\mb{w}_i$ to its closest representative. This sub-problem is described in lines $19$-$21$ of Table \ref{table:algorithm}. 

\begin{table}[t]
\begin{tabular}{l}
\hline
1: \textbf{procedure} DL-based sub-space clustering  \\
2: \textbf{Input}: original sub-vectors $\mb{W}$, \# of representatives $K_{dl}$\\
 \quad dictionary size $L_{dl}$, sparsity level $\alpha$.\\
3: Obtain initial solution $\{\mb{D}_0,\mb{\Lambda}_0,\mb{\Gamma}_0\}$\\
4: \textbf{repeat}\\
5: \quad\textbf{for} $i=1:K_{dl}$\quad//\textit{Sparse Coding}\\
6: \quad\textbf{Initialize}: $\mb{E}=\mb{W}_{I_i}$, $\mathcal{S}=\emptyset$\\
7: \quad\quad\textbf{for} $j=1:\alpha$\\
8: \quad\quad\quad Build new support: $\mathcal{S}\leftarrow \mathcal{S}\,\cup\, \{k\}$, \\ 
\quad\quad\quad\quad where $k=\arg\max_{j\not\in \mathcal{S}} |\mb{1}^{\textrm{T}}\mb{E}^{\textrm{T}}\mb{d}_j|$.\\ 
9: \quad\quad\quad Find new solution: $\bs{\zeta}_{\mathcal{S}}$ by solving $ \min_{\bs{\xi}} ||\mb{W}_{I_i}-\mb{D}_{\mathcal{S}}\,\bs{\xi}\,\mb{1}^\textrm{T}||_F^2$.\\
10:\quad\quad\quad Update residual: $\mb{E}=\mb{W}_{I_i}-\mb{D}_{\mathcal{S}}\,\bs{\zeta}_{\mathcal{S}}\,\mb{1}^\textrm{T}$.\\
11:\quad\quad\textbf{end} \\
12:\quad\textbf{end}\\
13:\quad\textbf{Initialize}: $\mb{E}=\mb{W}-\mb{D}\mb{\Lambda}\mb{\Gamma}$\quad //\textit{Dictionary Update}\\
14:\quad\textbf{for} $i=1:L_{dl}\quad$\\
15:\quad\quad Modify error: $\mb{F}=\mb{E}_{I_i}+\mb{d}_i\,\widetilde{\mb{g}}_{i,I_i}$.\\
16:\quad\quad Update $i$-th atom:    $\mb{d}_i=\frac{\mb{F}\,(\widetilde{\mb{g}}_{i,I_i})^{\textrm{T}}}{||\mb{F}\,(\widetilde{\mb{g}}_{i,I_i})^{\textrm{T}}||}$.\\
17:\quad\quad Re-compute error: $\mb{E}_{I_i}=\mb{F}-\mb{d}_i\,\widetilde{\mb{g}}_{i,I_i}$.\\
18:\quad\textbf{end}\\ 
19:\quad\textbf{for} $i=1:p^2M\quad$//\textit{Assignment Update}\\
20:\quad\quad Update $\bs{\gamma}_i$ solving $j_i=\arg\min_{j\in\{1,\dots,K_{dl}\}} ||\mb{w}_i-\widetilde{\mb{c}}_j||_2$.\\
21:\quad\textbf{end}\\ 
22:\textbf{Until}: a maximum number of iterations is met.\\
23:\textbf{Return}: $\mb{D}$, $\mb{\Lambda}$, $\mathbf{\Gamma}$.\\
24:\textbf{end procedure}\\
\hline
\end{tabular}
\caption{Proposed algorithm for solving \eqref{eq:ss_dl_dagnostic}}
\label{table:algorithm}
\end{table}

\subsection{Initial Solution and Parameter Selection}
In order to provide an initial solution $\mb{D}_0$, $\mb{\Lambda}_0$, $\mb{\Gamma}_0$, to the proposed acceleration technique, we work as follows:
\begin{enumerate}
    \item We obtain a clustering of the original kernel sub-vectors into $K_{dl}$ clusters by minimizing $||\mb{W}-\mb{C}\mb{\Gamma}||_F$
    under the constraints on the assignment matrix $\mb{\Gamma}$ stated in \eqref{eq:ss_dl_dagnostic}. This problem can be solved by using the $k$-means algorithm.
    \item We then obtain a sparse representation of the cluster centroids in $\mb{C}$ as $\mb{C}\approx \mb{D}_0\mb{\Lambda}_0$, 
    by using a dictionary of size $L_{dl}$ and target sparsity $\alpha$. This problem can be solved with standard DL techniques such as the ones described earlier. 
    \item Finally, we obtain the initial assignment matrix $\mb{\Gamma}_0$ by assigning each of the sub-vectors in $\mb{W}$ to its closest representative in $\widetilde{\mb{C}}=\mb{D}_0\mb{\Lambda}_0$.
    \end{enumerate}
There are four free parameters in the proposed technique, namely, the subspace dimension $N'$, the number of representatives $K_{dl}$, the size of the dictionary $L_{dl}$, and the sparsity level $\alpha$. 
For a target acceleration $\rho$, we first determine the number of representatives $K_{vq}$ required by the VQ technique in order to achieve $\rho$, by using \eqref{eq:accel_vq}. This provides a lower bound for the number of representatives $K_{dl}$ required by the proposed technique. We set $K_{dl}=c\, K_{vq}$, $c>1$. Then, for a target sparsity $\alpha$, we use \eqref{eq:T_dl vs T_vq} with equality in order to determine the dictionary size required to achieve $\rho$. Typical ranges for the parameter values are $c=2,\ldots,5$, $\alpha=1,2,3$, and $N'=4,\ldots,8$. 


\section{Experimental Results}\label{sec:simulations}

In this section, the performance of the proposed technique is evaluated and compared against the conventional VQ approach defined in \eqref{eq:W_approx_CG} (and used in \cite{Cheng2018_quantized}). A two-fold performance evaluation is presented here. Specifically, in Experiment I, we evaluate the representation power of the rivals by means of the achieved quantization error, namely the error between $\mb{W}$ and its approximations defined in \eqref{eq:W_approx_CG} and \eqref{eq:W_approx_DLG}, respectively, for a range of target accelerations. This experiment is performed on the basis of individual-layer kernel approximations.

As expected, the less the per-layer quantization error, the less the anticipated accuracy loss of the accelerated model. Measuring this loss is the topic of Experiment II, where we perform full-range acceleration for selected modern DNNs, and compare the achieved accuracy of the accelerated models. In this case, the acceleration is limited to conv layers which are responsible for the vast majority of the DNN's computational complexity.  

Our experiments are based on pre-trained versions of three state-of-the-art DNNs for image classification, namely, VGG-16 \cite{Simonyan2014}, SqueezeNet \cite{Iandola2016}, and ResNet18 \cite{He2016}, using the training and validation datasets of ILSVRC2012 \cite{Russakovsky2015}, for fine-tuning and accuracy evaluation purposes, respectively. 

\subsection{Experiment I. Quantization error in individual layers}

In the first experiment, we evaluate the quantization error of the proposed technique for a range of target accelerations and compare the results against the conventional, k-means-based, VQ technique. To this end, we approximate the kernels of individual convolutional layers from the VGG16, SqueezeNet, and ResNet18 networks, using \eqref{eq:W_approx_CG} and \eqref{eq:W_approx_DLG}, and measure, in each case, the mean error between the original and approximated weights. For the target acceleration ratios, the number of representatives $K_{vq}$ required by the VQ technique was calculated via \eqref{eq:T(F)_vq}. 
Then, by setting the number of DL representatives as $K_{dl}=c\,K_{vq}$, $c>1$, the dictionary size $L_{dl}$ was obtained so that \eqref{eq:T_dl vs T_vq} holds with equality. We then calculated the quantization error versus the achieved acceleration for various selections of the coefficient $c$ and the sparsity level $\alpha$. The subspace dimension was set to $N'=8$, which is a typical value used in the relevant bibliography. 

\begin{figure*}
    \centering
    \begin{minipage}[t]{0.3\linewidth}
     \centering
    {\includegraphics[width=.99\linewidth]{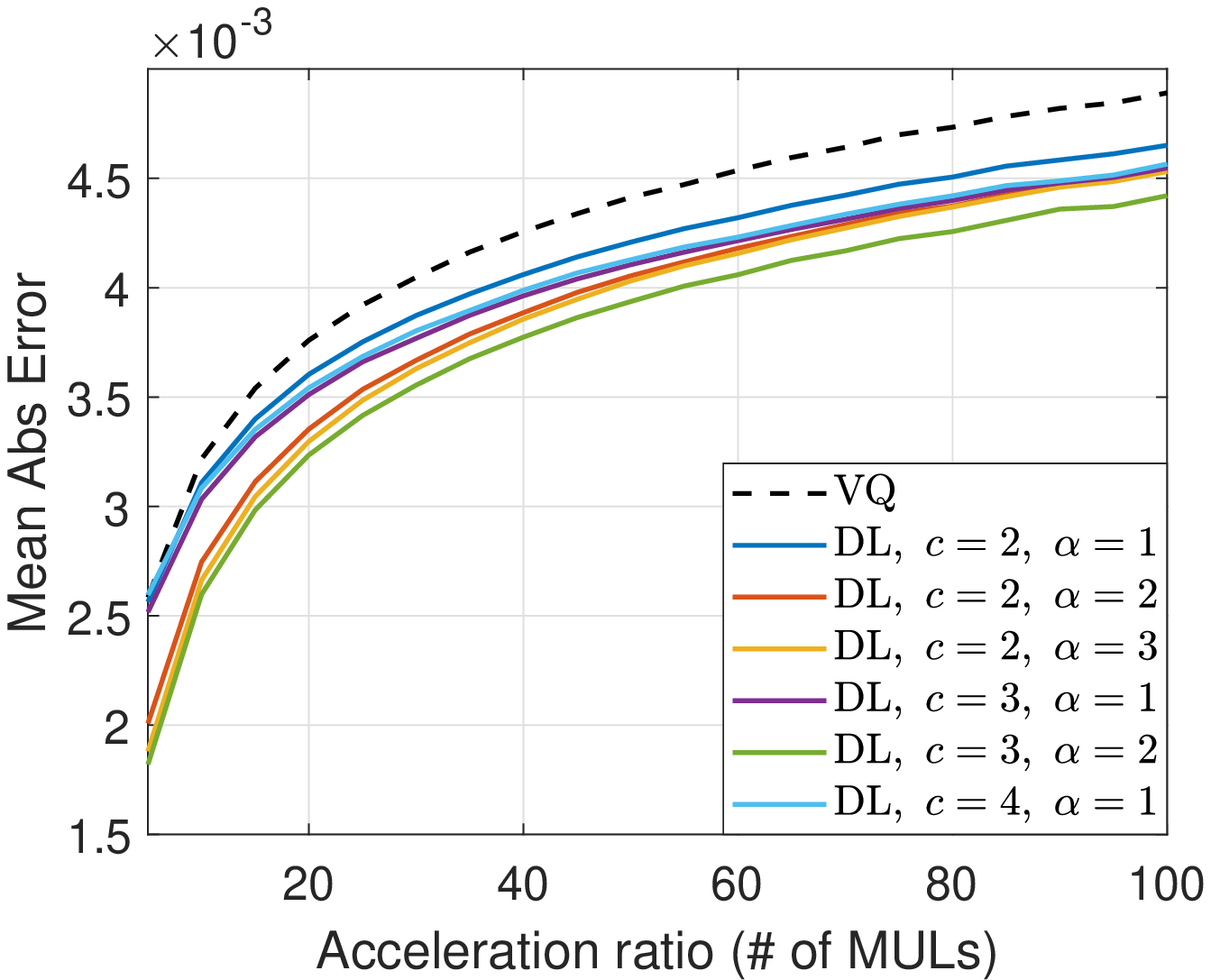}}
   \end{minipage}
    \begin{minipage}[t]{0.3\linewidth}
     \centering
    {\includegraphics[width=.99\linewidth]{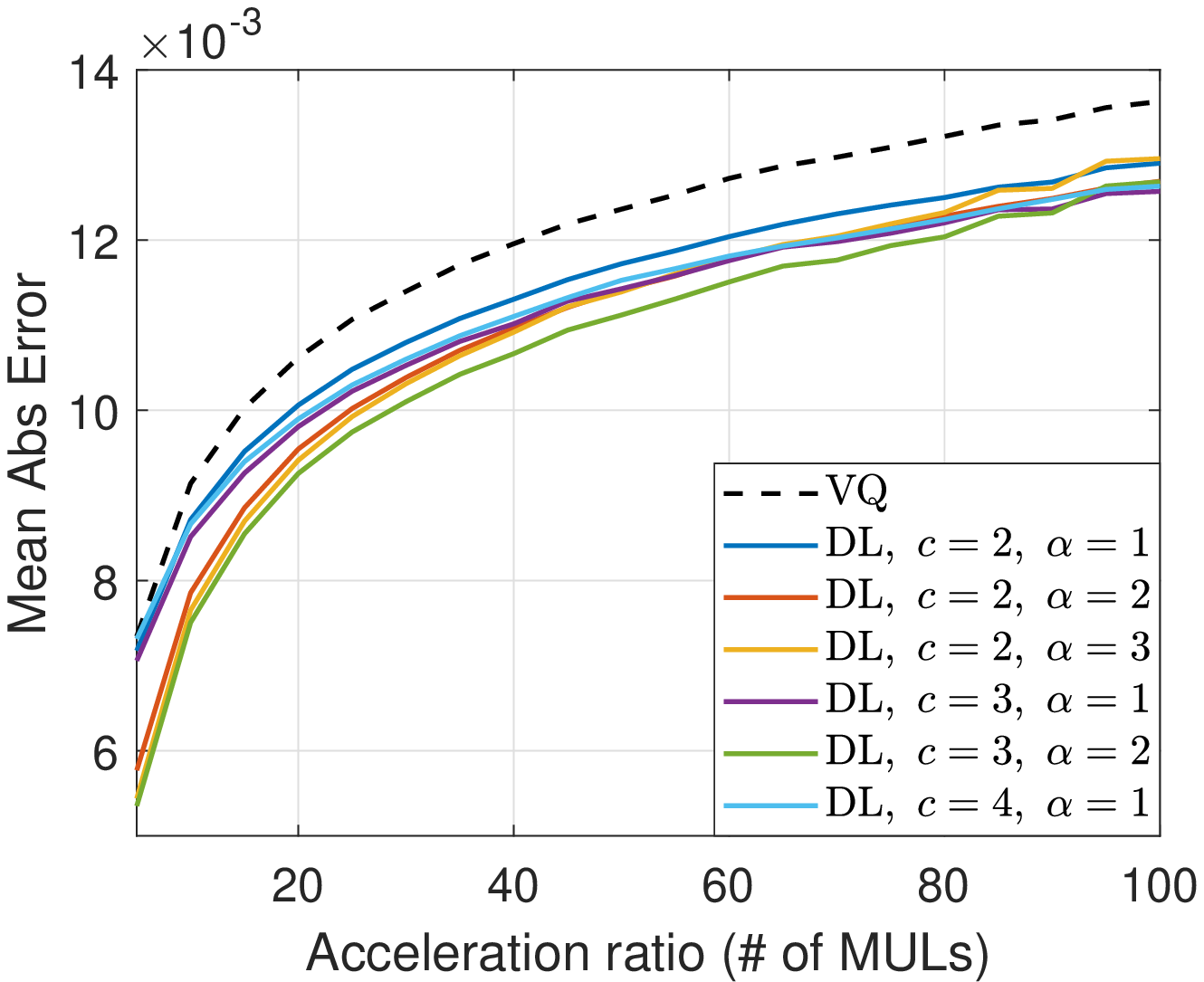}}
    \end{minipage}
    \begin{minipage}[t]{0.3\linewidth}
     \centering
    {\includegraphics[width=.99\linewidth]{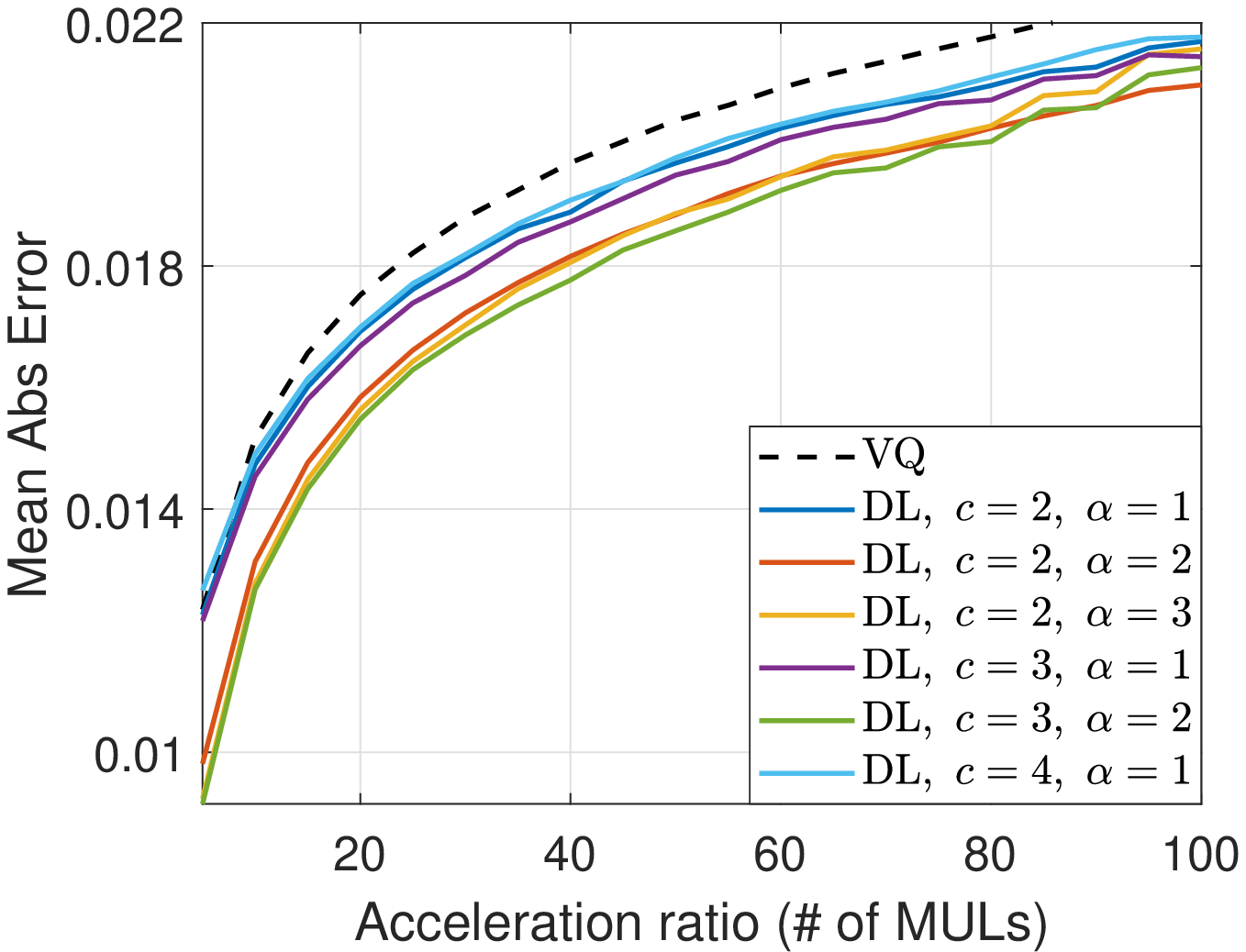}}
    \end{minipage}
    \\
    \begin{minipage}[t]{0.3\linewidth}
     \centering
    {\includegraphics[width=.99\linewidth]{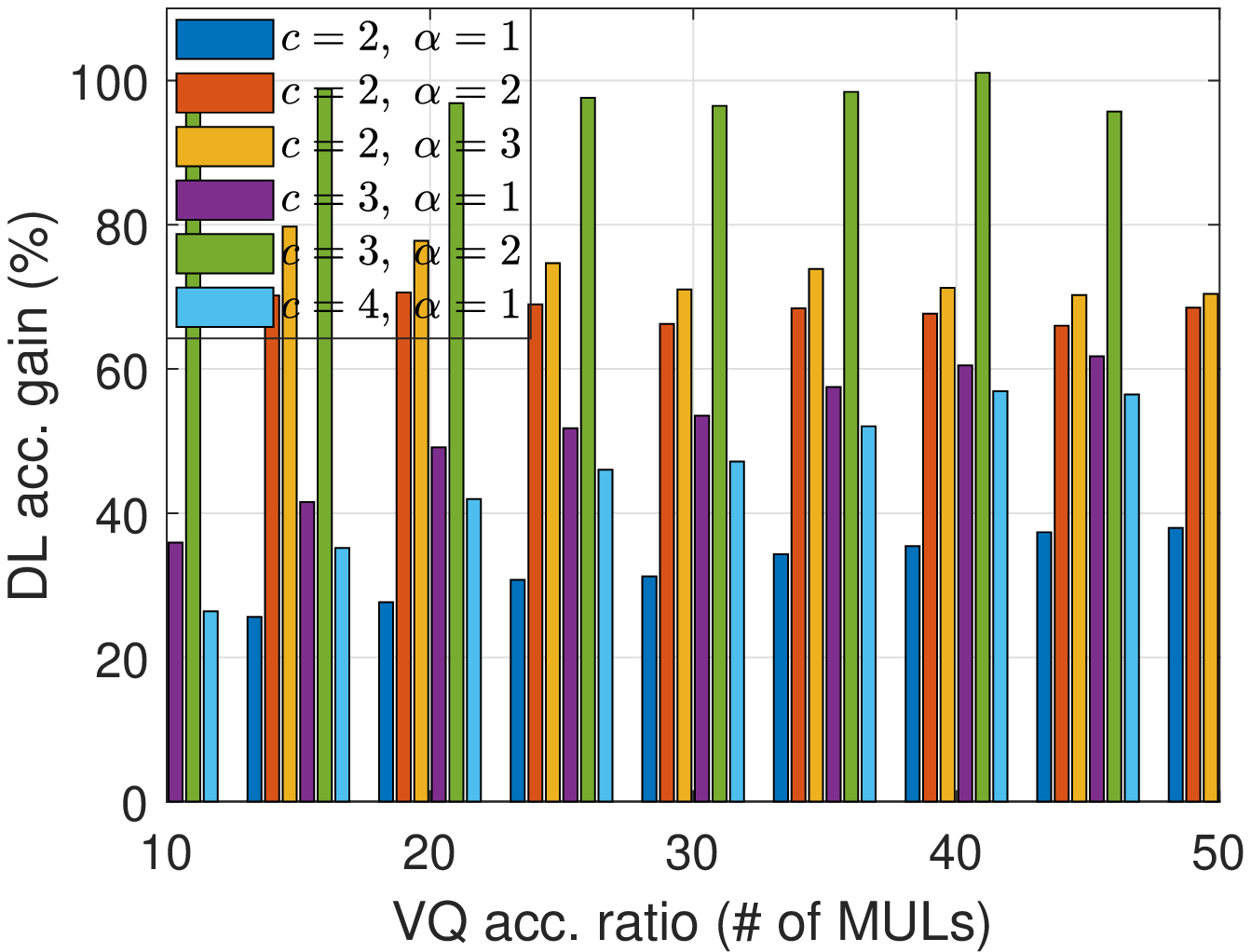}}
    \centerline{ \scriptsize{(a)}}\medskip
    \end{minipage}
     \begin{minipage}[t]{0.3\linewidth}
     \centering
    {\includegraphics[width=.99\linewidth]{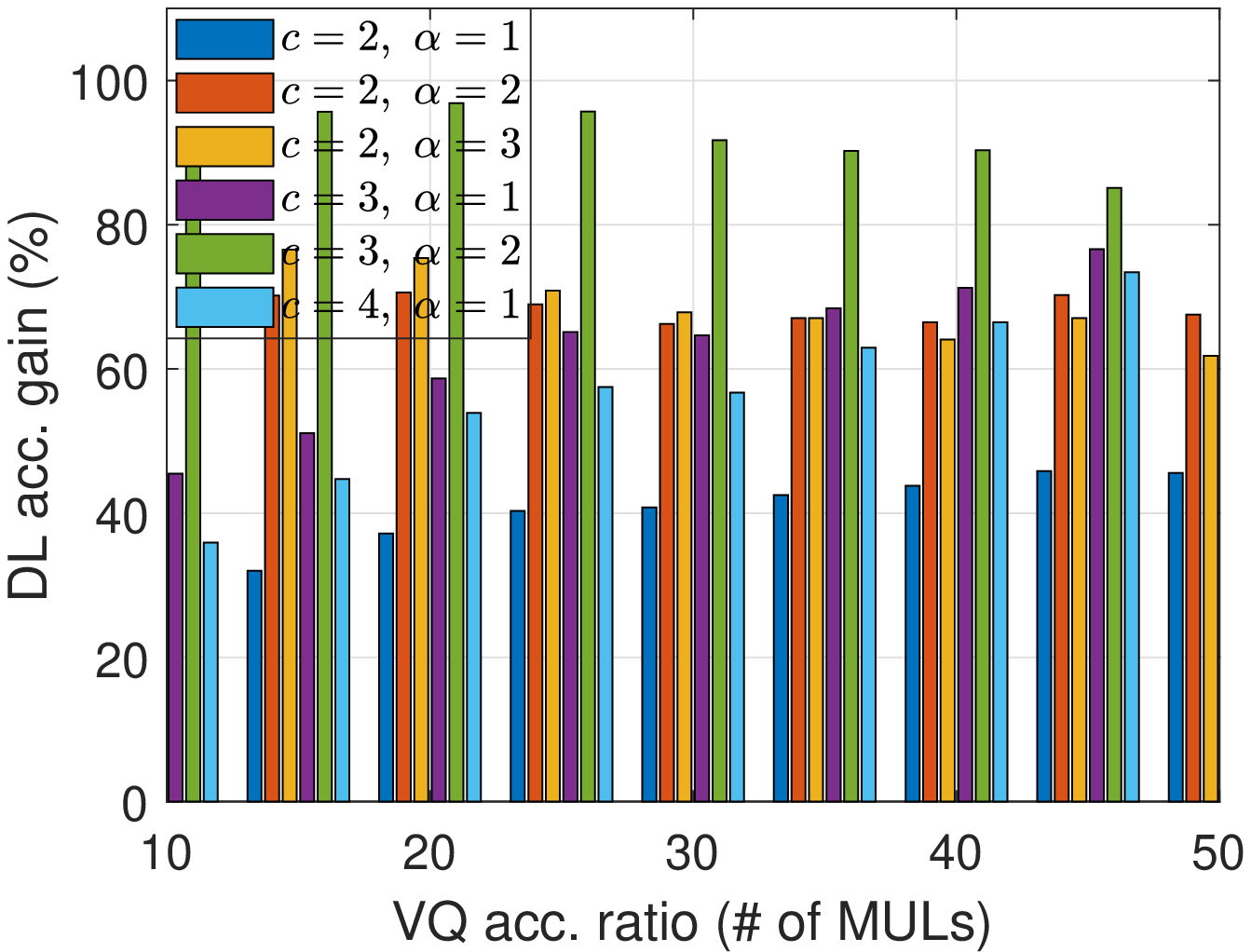}}
    \centerline{ \scriptsize{(b)}}\medskip
    \end{minipage}
    \begin{minipage}[t]{0.3\linewidth}
     \centering
    {\includegraphics[width=.99\linewidth]{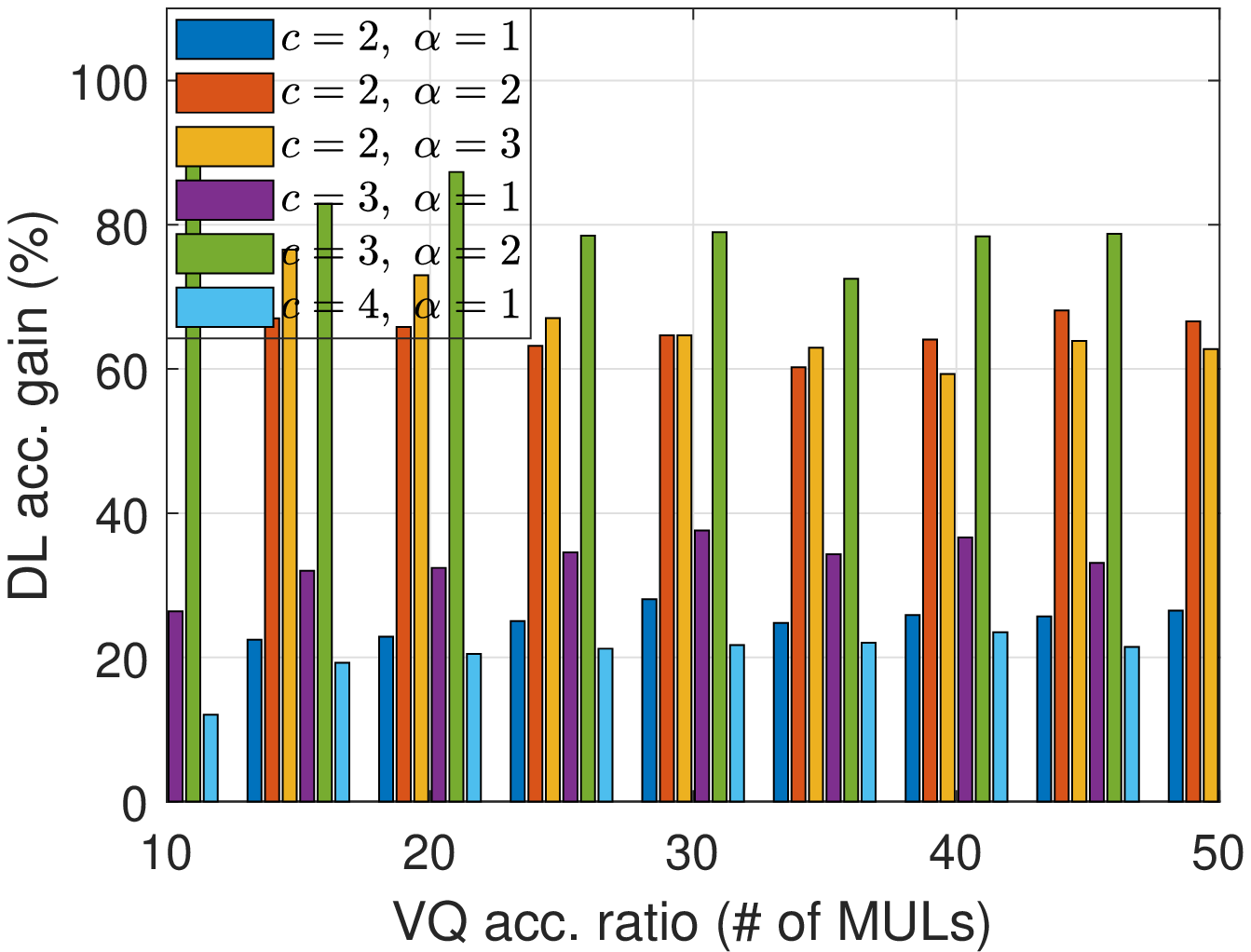}}
    \centerline{ \scriptsize{(c)}}\medskip
    \end{minipage}
    \vspace{-5pt}
    \caption{Mean quantization error (top row) and acceleration gain (bottom row) of DL vs VQ techniques as a function of the acceleration ratio for layers: (a) layer $\textrm{conv}4-1$ of VGG16, (b) layer $\textrm{res4a-branch2b}$ of ResNet18, and (c) layer $\textrm{fire8-expand3x3}$ of SquezeNet. In all cases, the subspace dimension was $N'=8$.}
    \vspace{-10pt}
    \label{fig:Exp_I}
\end{figure*}

A representative instance of Experiment I involving three selected conv layers from the used models, namely, (a) $\textrm{conv}4-1$ of VGG16 ($512$ kernels of size $3\times 3\times 256$), (b) $\textrm{res4a-branch2b}$ of ResNet18 ($256$ kernels of size $3\times 3\times 256$), and (c) $\textrm{fire8-expand3x3}$ of SquezeNet ($256$ kernels of size $3\times 3\times 64$), is shown in Fig.  \ref{fig:Exp_I} (top row). As shown in Fig. \ref{fig:Exp_I}, (and will become more apparent in the Experiment II), the proposed technique clearly outperforms its rival with respect to quantization error, achieving a better approximation of the original weights, for the same acceleration. Equivalently, this superior performance is translated into a significant acceleration gain for the same quantization error, as quantified by the respective plots on the bottom row of Fig. \ref{fig:Exp_I}.


\begin{figure*}
         \centering
        \begin{minipage}[t]{0.3\linewidth}
         \centering
        {\includegraphics[width=.99\linewidth]{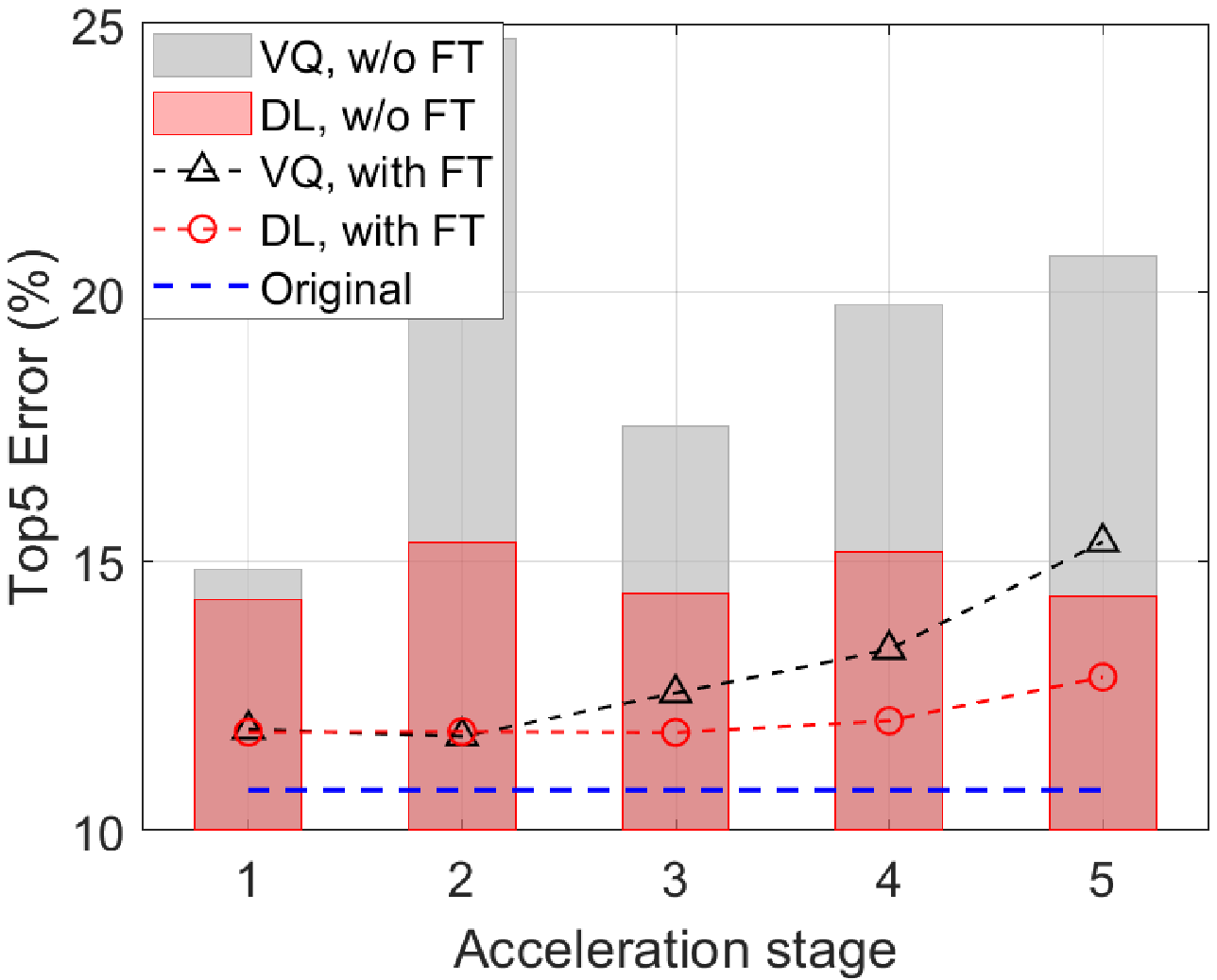}}
        \centerline{ \scriptsize{VGG16, $\rho=20$}}\medskip
        \end{minipage}
        \begin{minipage}[t]{0.3\linewidth}
         \centering
        {\includegraphics[width=.99\linewidth]{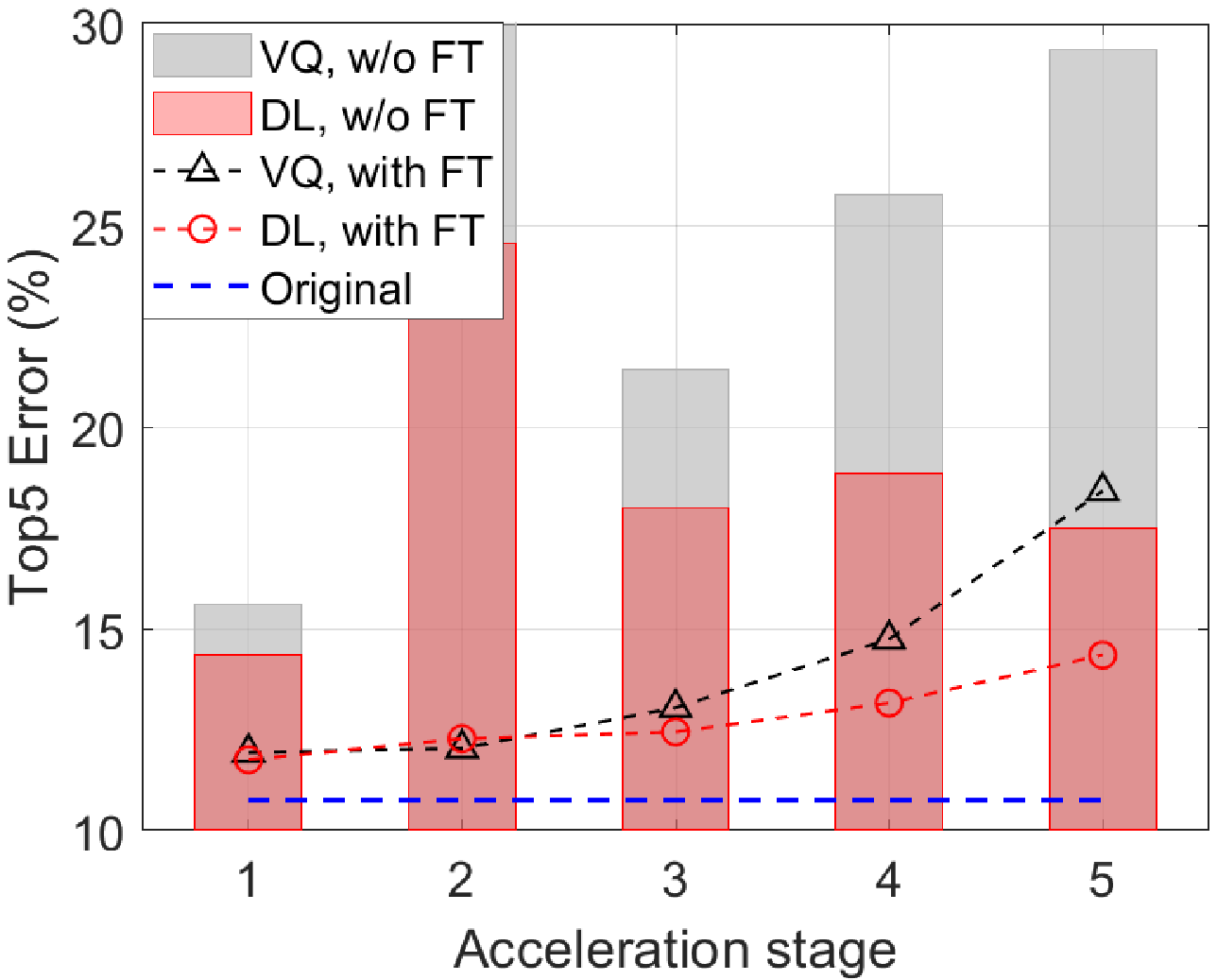}}
        \centerline{ \scriptsize{VGG16, $\rho=30$}}\medskip
        \end{minipage}
        \begin{minipage}[t]{0.3\linewidth}
         \centering
        {\includegraphics[width=.99\linewidth]{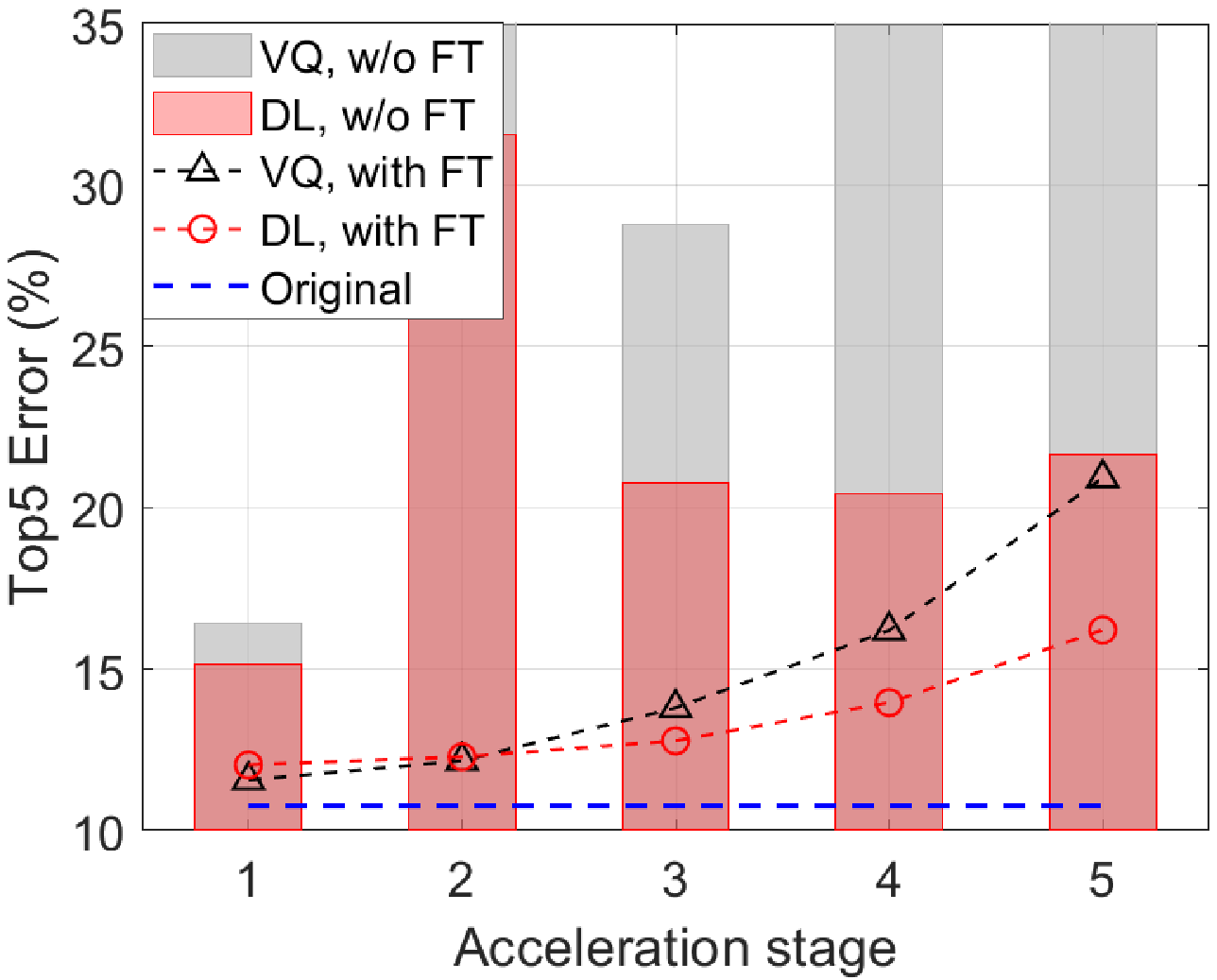}}
        \centerline{ \scriptsize{VGG16, $\rho=40$}}\medskip
        \end{minipage}
        \\
        \vspace{-3pt}
         \begin{minipage}[t]{0.3\linewidth}
         \centering
        {\includegraphics[width=.99\linewidth]{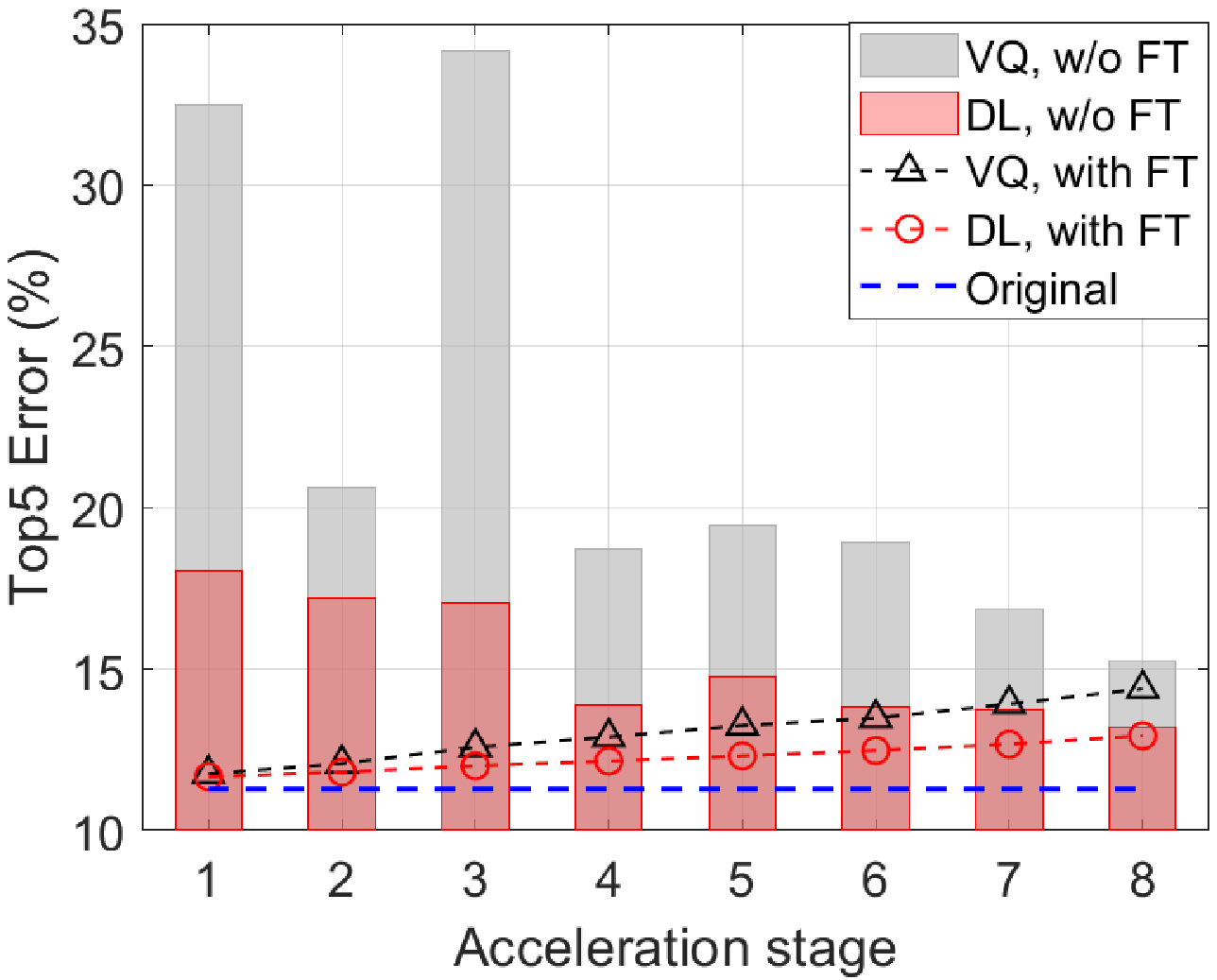}}
        \centerline{ \scriptsize{ResNet18, $\rho=10$}}\medskip
        \end{minipage}
        \begin{minipage}[t]{0.3\linewidth}
         \centering
        {\includegraphics[width=.99\linewidth]{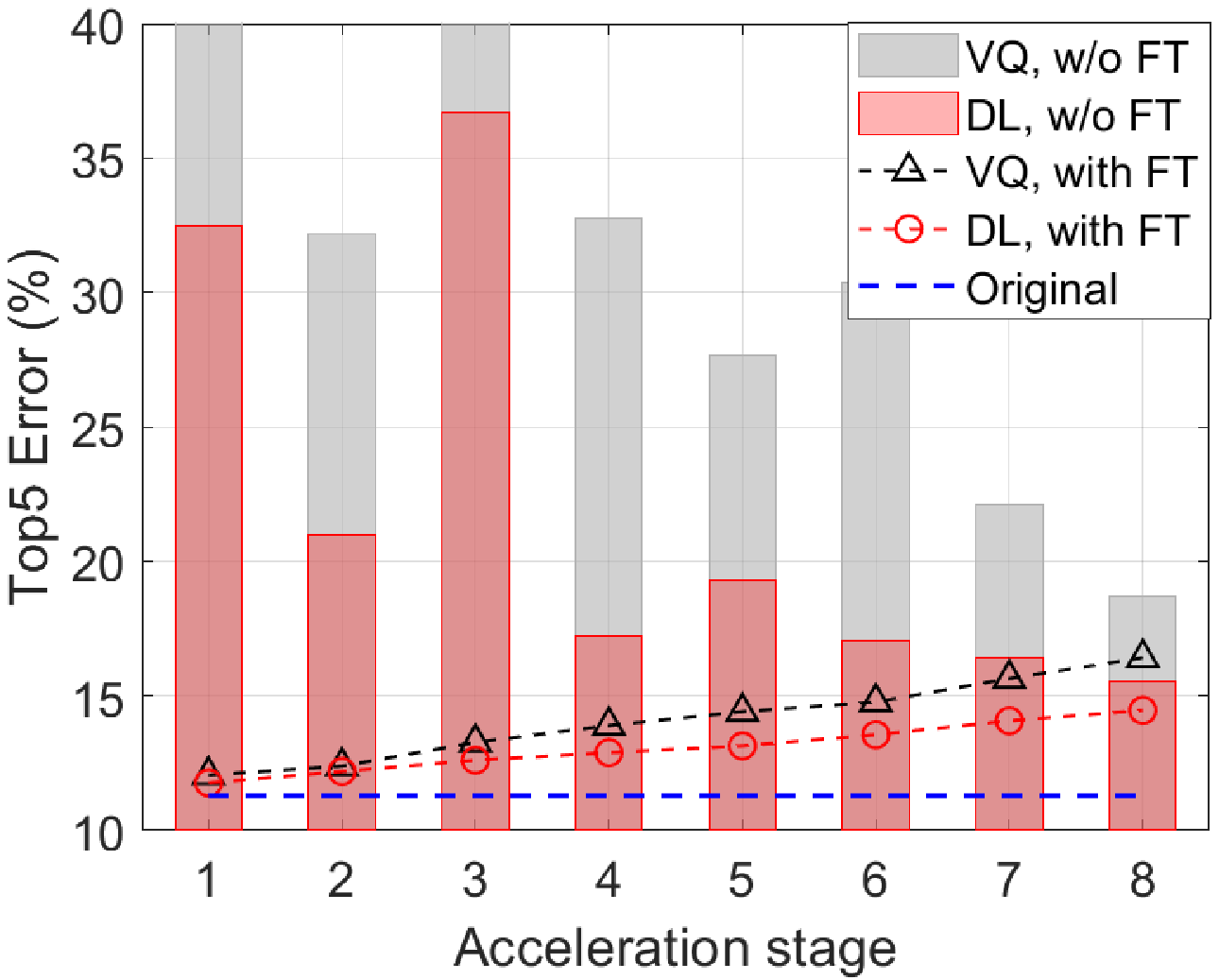}}
        \centerline{ \scriptsize{ResNet18, $\rho=20$}}\medskip
        \end{minipage}
        \begin{minipage}[t]{0.3\linewidth}
         \centering
        {\includegraphics[width=.99\linewidth]{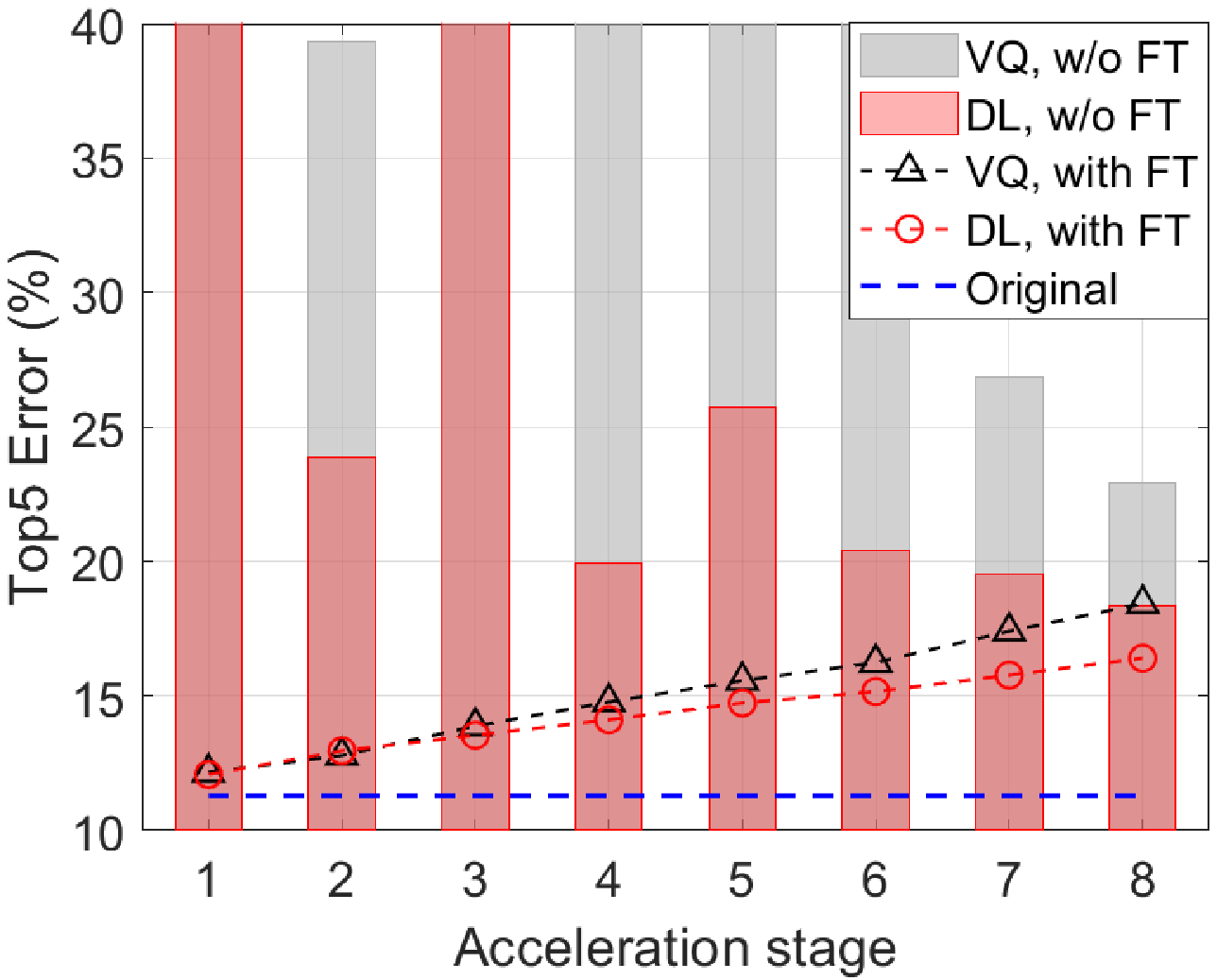}}
        \centerline{ \scriptsize{ResNet18, $\rho=30$}}\medskip
        \end{minipage}
        \\
        \vspace{-3pt}
        \begin{minipage}[t]{0.3\linewidth}
         \centering
        {\includegraphics[width=.99\linewidth]{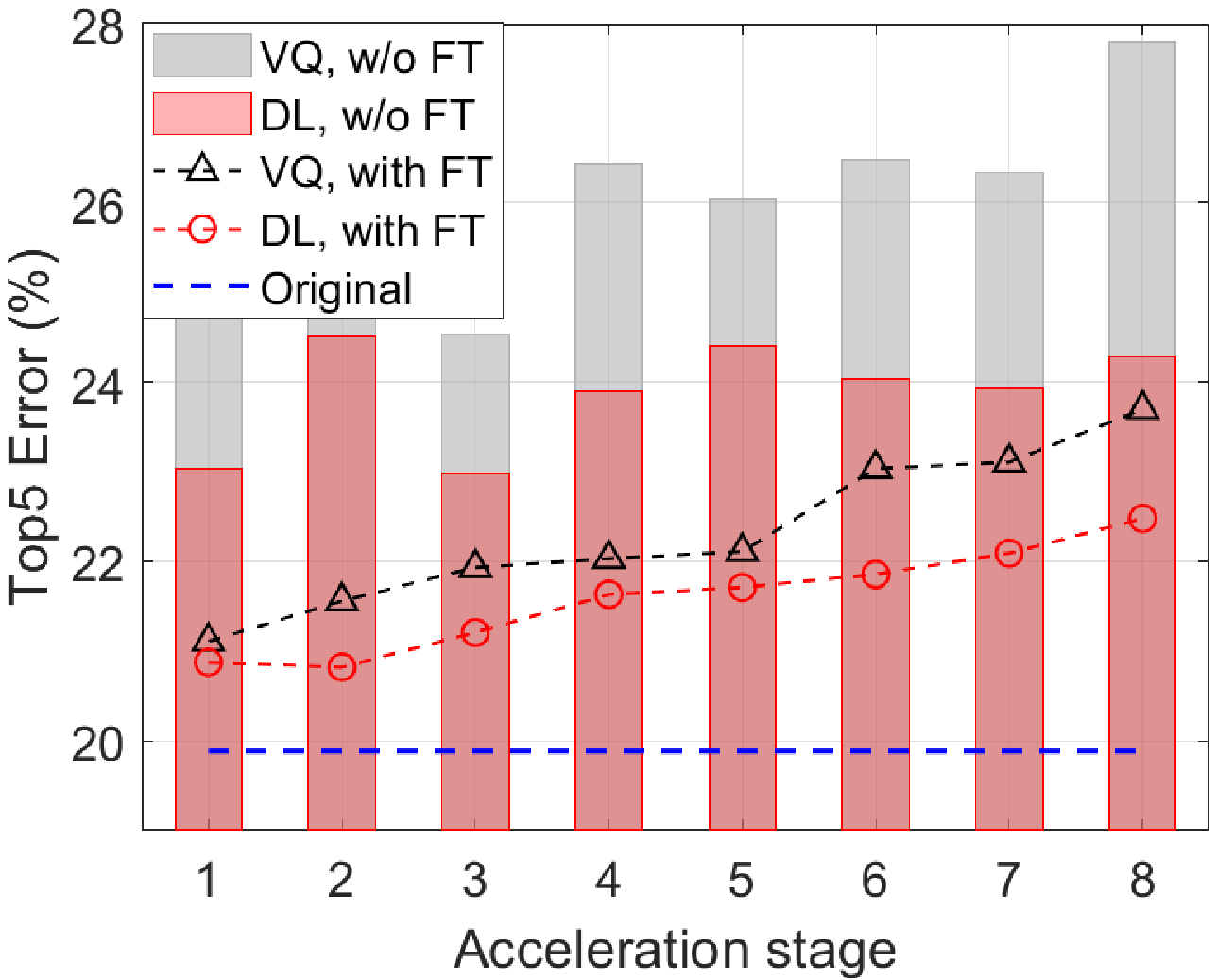}}
        \centerline{ \scriptsize{SqueezeNet, $\rho=10$}}\medskip
        \end{minipage}
        \begin{minipage}[t]{0.3\linewidth}
         \centering
        {\includegraphics[width=.99\linewidth]{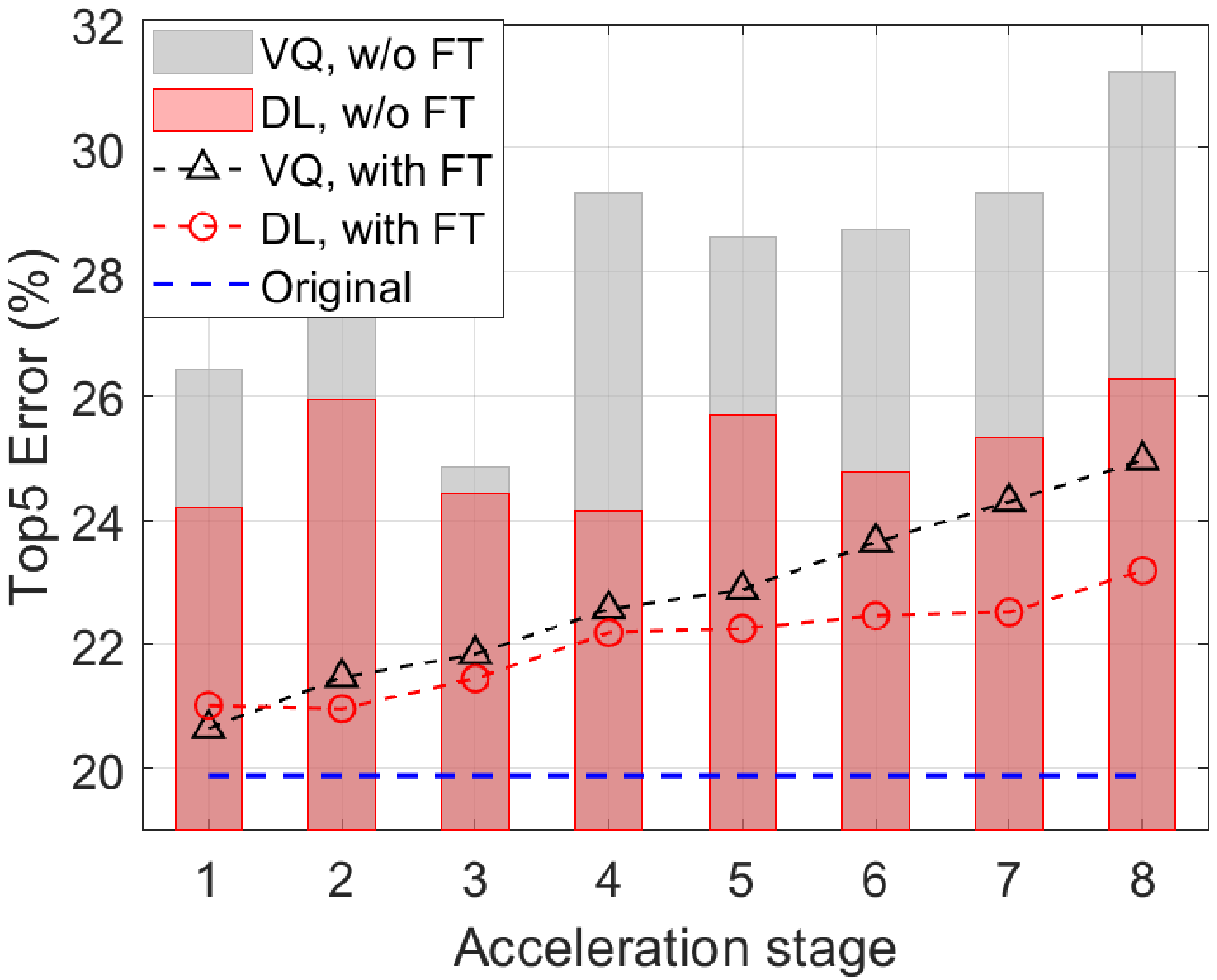}}
        \centerline{ \scriptsize{SqueezeNet, $\rho=15$}}\medskip
        \end{minipage}
        \begin{minipage}[t]{0.3\linewidth}
         \centering
        {\includegraphics[width=.99\linewidth]{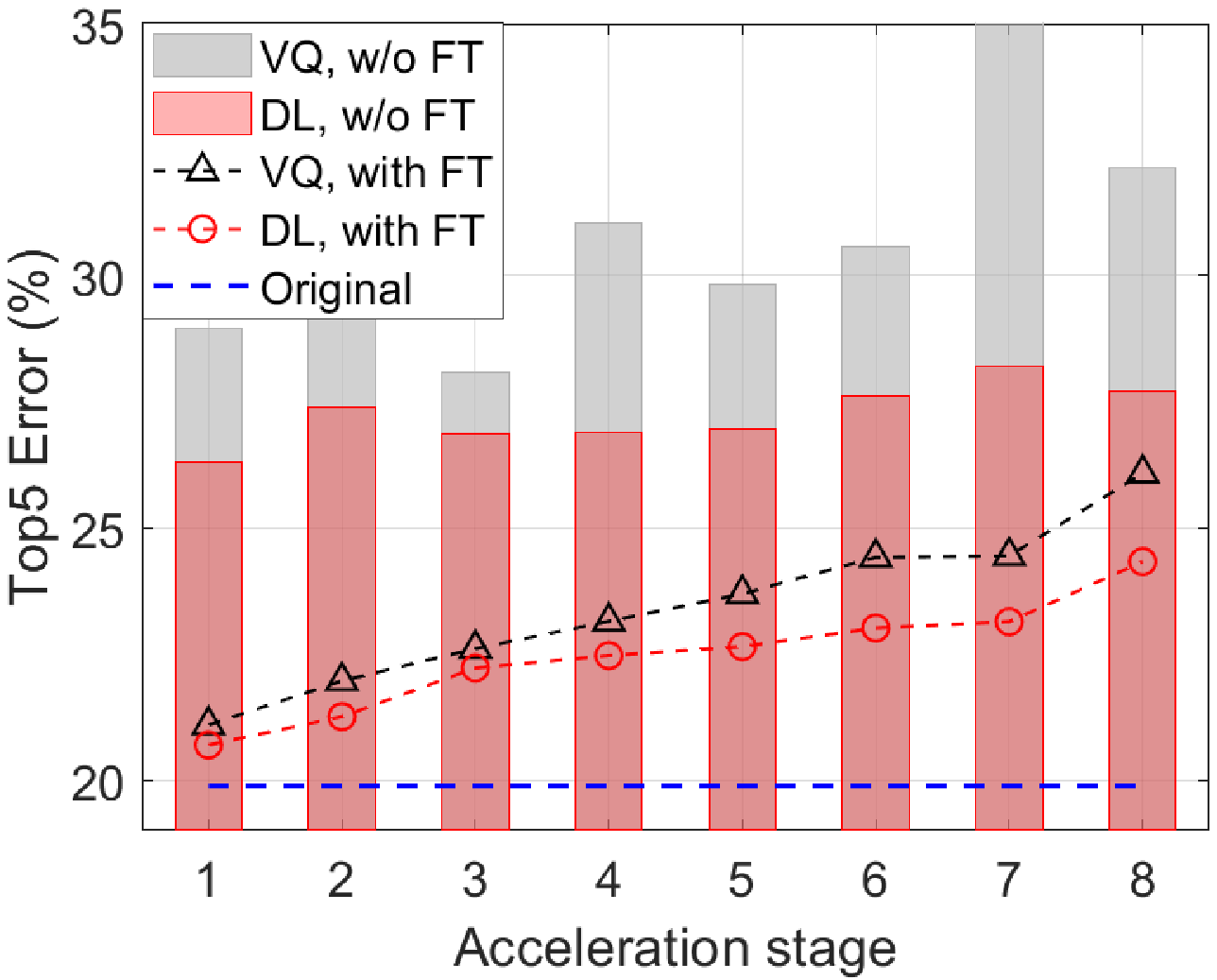}}
        \centerline{ \scriptsize{SqueezeNet, $\rho=20$}}\medskip
        \end{minipage}
        \vspace{-5pt}
    \caption{Full-model acceleration for the convolutional layers of VGG-16 (top), ResNet18 (middle), and SqueezeNet (bottom). The acceleration is performed sequentially in $5$, $8$, and $8$ stages, respectively. Each stage involves accelerating one or more conv layers and measuring the validation performance of the network before and after fine-tuning. In each case, the starting point for stage $i$ is the corresponding accelerated and fine-tuned network from stage $i-1$. The parameters values were $N'=8$, $c=3$, and $\alpha=2$, while the acceleration ratio $\rho$ was common for all stages, and it is reported on the bottom of the plots.}
    \label{fig:full_model_acc}
    \vspace{-10pt}
\end{figure*}

\subsection{Experiment II. Accuracy loss}

In this experiment, we apply the rival techniques to the three DNN models in a ``full-model'' acceleration scenario. It involves accelerating multiple (or all) convolutional layers of the original models and measuring the achieved classification accuracy of the accelerated networks. 

It is stressed here that, although full-range acceleration depends heavily on the performance of the technique used for the acceleration of each layer, it also involves experimentation over the strategy used for accelerating the layers and the involved fine-tuning (re-training) of the accelerated model. Here, we follow a stage-wise acceleration approach (as proposed in \cite{Cheng2018_quantized}) with each stage involving accelerating (and fixing) one or more layers of the network, and subsequently, fine-tuning (i.e., re-training) the remaining original layers. The starting point for each stage is the accelerated and fine-tuned version of the previous stage. The process begins with the original network, and it is repeated until all target layers are accelerated. For fine-tuning and performance assessment we use the training and validation datasets, respectively, from ILSVRC2012. Since, in each stage, only a small fraction of the network is affected and in order to expedite the process, we divided the initial training dataset into smaller subsets and used these smaller sets for fine-tuning purposes.

\paragraph{Accelerating VGG16}

VGG16 consists of $13$ $3\times 3$ convolutional and $3$ fully-connected (fc) layers and it is (by far) the most computationally intensive network of the three used in our experiments. Of the total $15.5\times 10^9$ MACs/MULs required for inference, over $99\%$ are consumed by the conv layers, meaning that the acceleration of the conv layers is practically equivalent to the acceleration of the entire network. The conv layers of VGG16 are organized in $5$ groups of consecutive conv layers, which is why we accelerate VGG16 in $5$ stages, with the $i$-th group being accelerated at stage $i$. The stage-by-stage results of our VGG16 acceleration experiment, using the procedure describe in the previous paragraph for three different acceleration ratios, namely $\rho=20$, $\rho=30$, and $\rho=40$, are shown in the top row of Fig. \ref{fig:full_model_acc}.

\paragraph{Accelerating ResNet18}
ResNet18 is based on the concept of residual learning and follows the architecture of other bigger ResNet variants (e.g. ResNet34, ResNet50, etc.). Its building block comprises of two consecutive $3\times 3$ conv layers, with the block's output being summed to its input using a ``bypass'' connection (hence, the block is required to only learn the residual representation). ResNet18 consists of $8$ such blocks (plus an input conv layer and an fc layer), that, similarly to the VGG16 case, are responsible for roughly $99\%$ of the total $1.8\times 10^9$ MACs/MULs required by the network. In our experiments with ResNet18 we accelerated its building blocks in a one-block-per-stage fashion leading to $8$ total acceleration stages. The acceleration results using ratios $\rho=10$, $\rho=20$, and $\rho=30$, are shown in the middle row of Fig. \ref{fig:full_model_acc}.

\paragraph{Accelerating SqueezeNet}
SqueezeNet is a fully convolutional CNN that employs a special architecture managing to drastically reduce its size while still remaining within the state-of-the-art performance territory. Its building block is the ``fire'' module that consists of a ``squeeze'' $1\times 1$ conv layer with the purpose of reducing the number of input channels, followed by   $1\times 1$ and $3\times 3$ ``expand'' conv layers that are connected in parallel to the ``squeezed'' output. SqueezeNet consists of $8$ such modules, connected in series. Since SqueezeNet constitutes an already ``streamlined'' network, in our acceleration experiments we followed a moderate acceleration strategy only targeting the $3\times 3$ ``expand'' layers that are responsible for roughly $53\%$ of the total $3.9\times 10^8$ MACs/MULs required by the network. Acceleration was performed in a one-module-per-stage fashion for a total of $8$ acceleration stages. The results for acceleration ratios $\rho=10$, $\rho=15$, and $\rho=20$, corresponding to a total acceleration of the network by $91\%$, $98\%$, and $101\%$, respectively, are shown in the bottom row of Fig. \ref{fig:full_model_acc}. 

As a general comment regarding the result presented in Fig. \ref{fig:full_model_acc}, it could be stated that the ``before fine-tuning'' error values (shown in bars) at each stage, reveal the sensitivity of the network with respect to accelerating/approximating the kernels of the layers involved at that particular stage, but also, the performance of the technique used to achieve the acceleration. As such, it is evident by the shown plots, that the proposed technique achieves a universally superior performance compared to its rival.

On the other hand, the corresponding ``after fine-tuning'' error values reflect the capacity of the remaining (original) layers to ``adapt'' to the newly accelerated part. Here we see that, by offering a better starting point to the fine-tuning process, the proposed technique still manages to outperform its rival by a safe margin in all cases, although, as it is to be expected, fine-tuning compresses the difference between the compared techniques to a great extent. 

In summary, both the comparative analysis of the results shown in Fig. \ref{fig:full_model_acc}, and also, the final Top5 error figures achieved by the accelerated networks, reveal a very promising performance by the proposed technique, whose application results in significantly accelerated CNNs, with limited loss of their classification power. It should be finally noted that the shown results could be further improved by following a more targeted acceleration strategy (e.g. experimentation over the acceleration sequence, the acceleration ratio per layer, using a more extensive fine-tuning process, etc.), which acts as further confirmation of our conclusion.

\section{Conclusions}
\label{sec:conclusions}
A new clustering-based weight-approximation technique for the acceleration of DNNs was proposed in this paper. The technique exploits the particularities of the problem at hand in order to increase the number of used centroids for the same target acceleration, as compared to the conventional $k$-means technique. This is achieved using a Dictionary Learning framework, by imposing a special structure to the centroids that reduces the overall computational complexity of the accelerated layer. The superior performance of the technique was validated via a number of experiments on three well-known state-of-the-art pre-trained DNN models. 

\section*{Acknowledgement}
This paper has received funding from H2020 project CPSoSaware (No 873718) and the DEEP-EVIoT - Deep embedded vision using sparse convolutional neural networks project (MIS 5038640) implemented under the Action for the Strategic Development on the Research and Technological Sector, co-financed by national funds through the Operational programme of Western Greece 2014-2020 and European Union funds (European Regional Development Fund).

\label{sec:ref}
\bibliographystyle{IEEEtran}
\bibliography{refs}

\end{document}